\documentclass[letterpaper, 10 pt, conference]{ieeeconf}  % Comment this line out if you need a4paper
\usepackage{graphicx} 
\usepackage{booktabs}
\usepackage{url} 
\IEEEoverridecommandlockouts                              % This command is only needed if 
\title{\LARGE \bf
Spatial-LLaVA: Enhancing Large Language Models with Spatial Referring Expressions for Visual Understanding
}

\author{Xuefei Sun$^{1}$, Doncey Albin$^{1}$, Cecilia Mauceri$^{2}$,Dusty Woods$^{1}$, and Christoffer Heckman$^{1}$% <-this % stops a space
\thanks{$^{1}$Authors are with the Autonomous Robotics and Perception Group in the Computer Science Department at the University of Colorado Boulder, Boulder, CO 80309, USA}%
\thanks{$^{2}$This work is supported by USDA-NIFA Award Number 2021-67021-33450 and by Chameleon.}%
}

\begin{document}

\maketitle
\thispagestyle{empty}
\pagestyle{empty}

%%%%%%%%%%%%%%%%%%%%%%%%%%%%%%%%%%%%%%%%%%%%%%%%%%%%%%%%%%%%%%%%%%%%%%%%%%%%%%%%
\begin{abstract}
Multimodal large language models (MLLMs) have demonstrated remarkable abilities in comprehending visual input alongside text input. Typically, these models are trained on extensive data sourced from the internet, which are sufficient for general tasks such as scene understanding and question answering. However, they often underperform on specialized tasks where online data is scarce, such as determining spatial relationships between objects or localizing unique target objects within a group of objects sharing similar features. In response to this challenge, we introduce the SUN-Spot v2.0 dataset\footnote{Project page: https://arpg.github.io/sunspot/}, now comprising a total of 90k image-caption pairs and additional annotations on the landmark objects. Each image-caption pair utilizes Set-of-Marks prompting as an additional indicator, mapping each landmark object in the image to the corresponding object mentioned in the caption. Furthermore, we present Spatial-LLaVA, an MLLM trained on conversational data generated by a state-of-the-art language model using the SUN-Spot v2.0 dataset. Our approach ensures a robust alignment between the objects in the images and their corresponding object mentions in the captions, enabling our model to learn spatial referring expressions without bias from the semantic information of the objects. Spatial-LLaVA outperforms previous methods by 3.15\% on the zero-shot Visual Spatial Reasoning benchmark dataset. Spatial-LLaVA is specifically designed to precisely understand spatial referring expressions, making it highly applicable for tasks in real-world scenarios such as autonomous navigation and interactive robotics, where precise object recognition is critical.
\end{abstract}

\section{Introduction}

\label{sec:intro}
Humans routinely refer to objects using their relative locations to other objects. These types of descriptions, known as spatial referring expressions, are vital for clarifying ambiguous instructions and distinguishing unique objects in cluttered environments. Humans excel at understanding these descriptions but despite recent advancements in text understanding and multimodal processing through foundation models, state-of-the-art models still struggle to reason over spatial concepts~\cite{kamath2023s}. Such descriptions are critical for tasks in the real world such as self-driving cars, and human robot interaction.
\begin{figure}[h]
\centering
\includegraphics[width=0.5\textwidth]{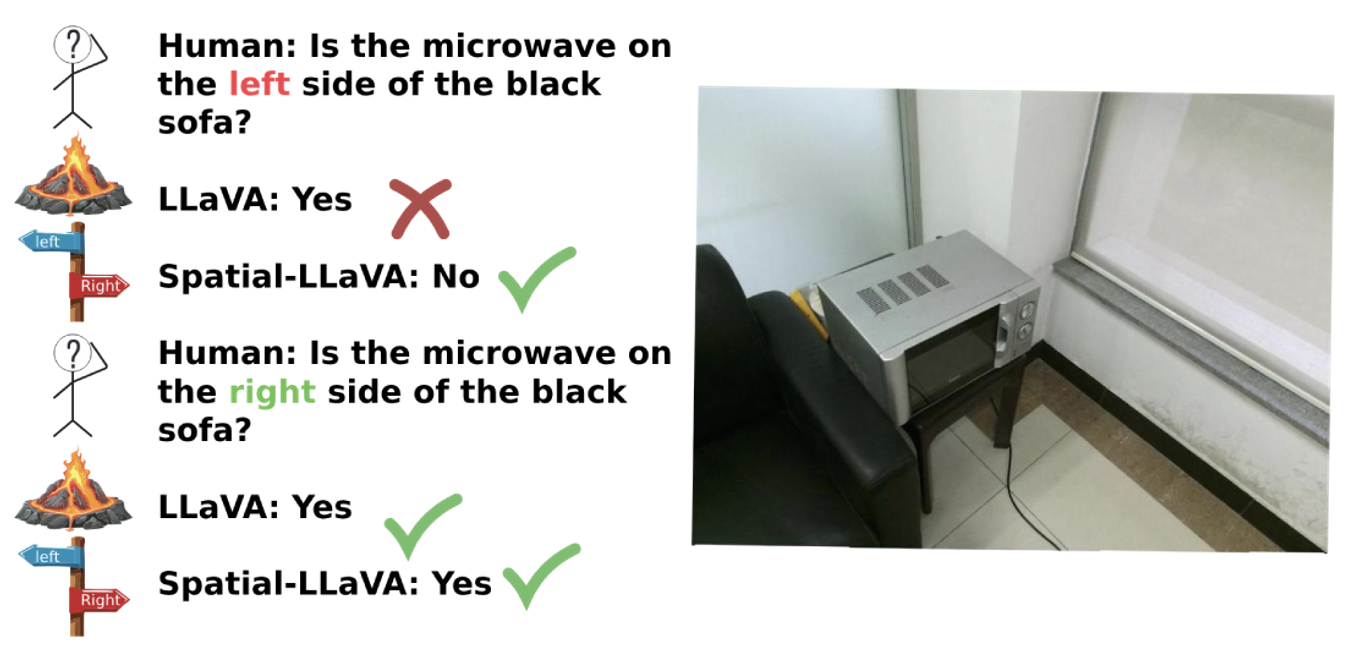}
\caption{While LLaVA Agrees With Both Queries, Spatial-LLaVA Accurately Identifies Right Side.}
\label{fig:motivation_example}
\end{figure}
Large Language Models (LLMs) are advanced AI systems adept at processing and generating human-like text. Rooted in complex neural network architectures such as transformers~\cite{vaswani2023attentionneed}, these models learn from vast datasets to grasp context and output natural language text responses, from answering questions~\cite{NEURIPS2023_9cb2a749} to storytelling~\cite{dettmers2023qloraefficientfinetuningquantized}. Despite their remarkable performance on many downstream tasks, LLMs face challenges when precise identification of specific targets within images is required~\cite{kamath2023s}. Most state-of-the-art LLMs are trained on the entire image and text describing events and main object in the image~\cite{chowdhery_2022_palm,touvron_2023_llama,geminiteam2024geminifamilyhighlycapable,openai_2023_gpt4}, which restricts their ability to focus on specific target objects and understand the relationships between these targets and their surrounding environment or landmarks~\cite{liu2023mmbench}.

Despite the wide applicability of spatial referring expressions, the available datasets~\cite{kazemzadeh2014referitgame,yu2016modeling,mao2016generationcomprehensionunambiguousobject,liu2019clevr,liu2021refer} focusing on spatial understanding provide limited annotations of landmarks present in the images which leads to suboptimal performance by state-of-the-art LLMs for tasks that require spatial awareness, rather than merely identifying object attributes like color, shape, and size~\cite{kamath2023s}.

Our work directly addresses these challenges through the introduction of %To address these challenges, we first introduce our 
SUN-Spot v2.0 dataset, the only RGB-D dataset that includes both spatial referring expressions and annotations on the landmark objects referred to in each corresponding expression. Compared to SUN-Spot dataset\cite{mauceri2019sun}, We enhanced the dataset by manually annotate each image with the landmark objects mentioned in the captions, ensuring precise identification of the referred objects. Further, We fine-tuned Spatial-LLaVA, a large multimodal language model designed to understand spatial relationships and distinguish objects within similar groups. Unlike prior methods trained on image-caption pairs~\cite{liu2023improvedLLaVA,dai2023instructblipgeneralpurposevisionlanguagemodels}, we adopt Set-of-Marks (SoM) prompting~\cite{yang2023setofmarkpromptingunleashesextraordinary}---a simple yet effective method that not only ensures object grounding in the captions but also enhances the learning of spatial relationships independent of object semantic information. Our contributions are summarized as follows:
\begin{itemize}
    \item We introduce the SUN-Spot v2.0 dataset, the first and only RGB-D dataset that includes spatial referring expressions and annotations for both target and landmark objects. The dataset features 90k spatial referring expressions across 10k images, averaging 9 captions per image.
    \item We present Spatial-LLaVA, a fine-tuned MLLM designed to predict spatial relationships between objects in images. 
    \item We integrate set-of-marks prompting into the fine-tuning pipeline to align objects in the images with their mentions in captions, offering a new perspective not commonly explored in previous work.
\end{itemize}

\section{Related Work}
\subsection{Multimodal Large Language Models}
%The field of artificial intelligence has experienced significant advancements
Significant advancements in natural language processing have been achieved with the advent of the Transformer architecture~\cite{devlin_2018_bert,liu_2019_roberta,yang_2019_xlnet}. This innovation, coupled with the realization that scaling model size and increasing data size often enhance performance on downstream tasks, has given rise to Large Language Models (LLMs) such as PaLM~\cite{chowdhery_2022_palm}, LLaMA~\cite{touvron_2023_llama}, Gemini~\cite{geminiteam2024geminifamilyhighlycapable} and GPT-4~\cite{openai_2023_gpt4}. These models demonstrate exceptional proficiency in text-based tasks, including speech recognition~\cite{journaloflatexclass_2015_mmger,heakl_6th,ma_an}, machine translation~\cite{heakl_6th,zhang_2023_prompting,moslem_2023_domain}, and information retrieval~\cite{labruna_2024_when,tang_selfretrieval}. 

Nevertheless, these models are limited to processing purely linguistic tasks. In real-world applications, multiple modalities, such as images, audio, and video, are frequently encountered. Consequently, researchers have turned their attention to leveraging the capabilities of LLMs to reason over inputs from various modalities. A typical Multimodal Large Language Model (MLLM) comprises three primary components: a modality-specific encoder~\cite{fang_2022_eva,radford_2021_learning,sun_2023_evaclip,mehdicherti_2023_reproducible}, a pretrained LLM~\cite{chung_2022_scaling,vicuna2023,touvron_2023_llama}, and a learnable connector. Encoders are selected based on the specific modality to extract modality-specific features~\cite{radford_2021_learning,sun_2023_evaclip}. The LLM is generally pre-trained on extensive language data. The learnable connector's function is to process and integrate features from different modalities, enabling them to be effectively input into the pre-trained LLM~\cite{liu2023improvedLLaVA,liu2023LLaVA,carion_2020_endtoend}.
The combination of these components allows MLLMs to process and interpret multimodal inputs, making them versatile and powerful tools in various real-world applications.

%For instance, in autonomous driving, MLLMs can simultaneously analyze visual inputs from cameras and commands/guidance from driver to make driving decisions~\cite{huang2024drivlmeenhancingllmbasedautonomous,chen2023drivingllmsfusingobjectlevel}. In healthcare, they can integrate medical imaging, patient history, and clinical notes to assist in diagnostics and treatment planning~\cite{choudhury2020role,li2024LLaVA,wang2024jmlr}. Moreover, the ability of MLLMs to handle diverse data types opens up new possibilities in human-computer interaction, enabling more natural and intuitive communication with AI systems.

\begin{figure}[]
\centering
\includegraphics[width=0.45\textwidth]{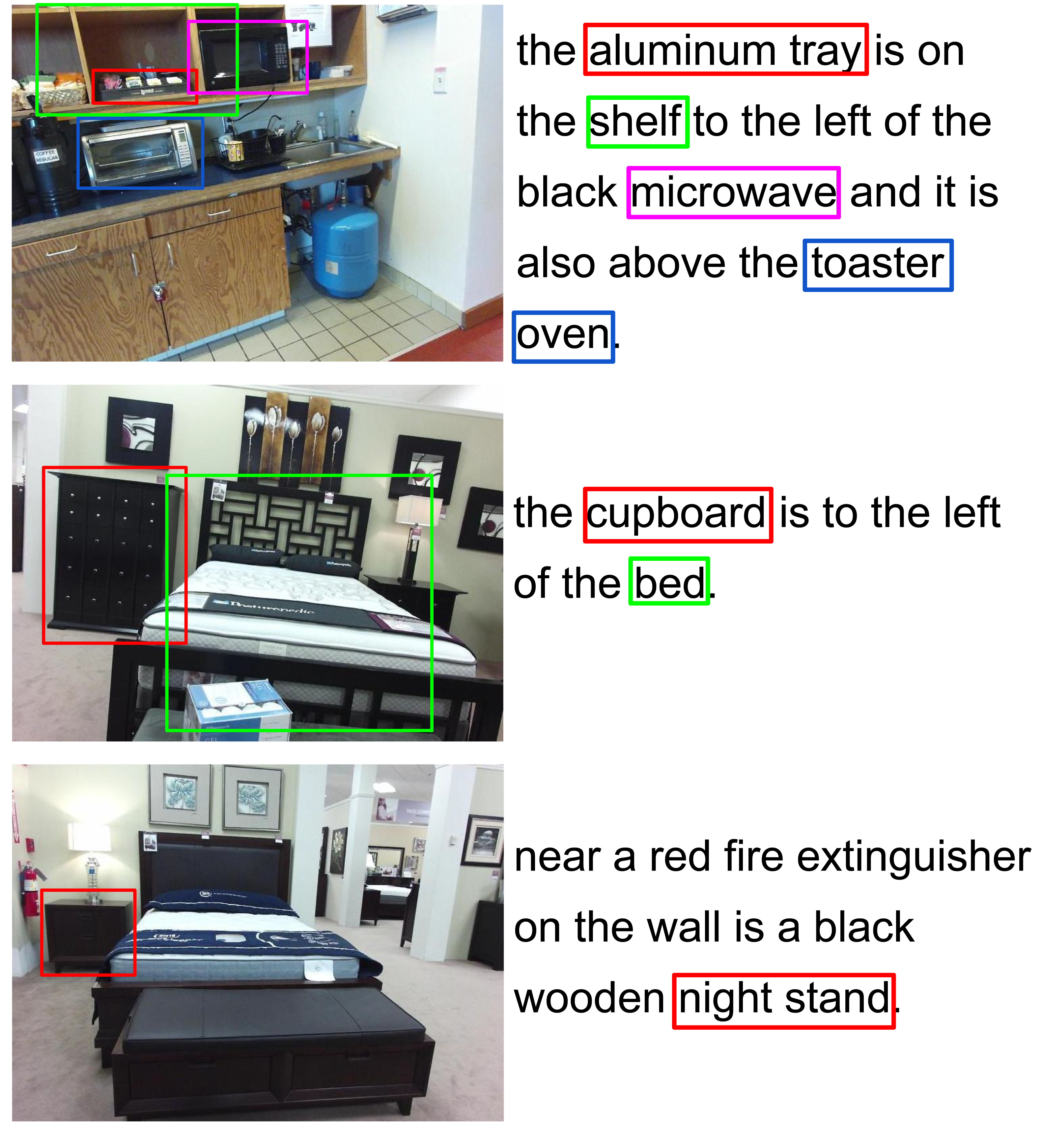}
\caption{Top: SUN-Spot Expert, Middle: SUN-Spot Machine-generated, Bottom: SUNRefer. Bounding boxes indicate landmarks are aligned to mentions.}
\label{fig:example_3_datasets}
\end{figure}

\begin{figure*}[]
\centering
\includegraphics[width=1\textwidth]{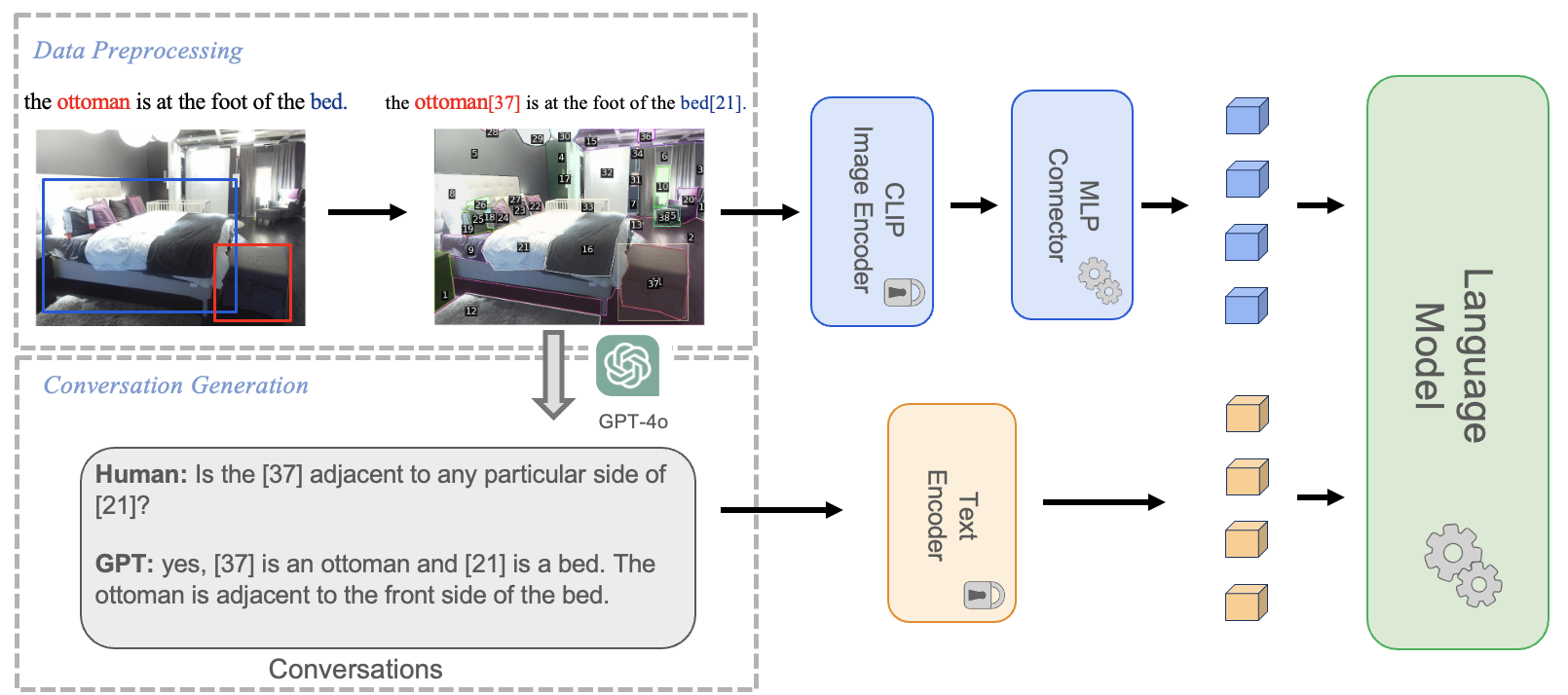}
\caption{System Structure}
\label{fig:system_example}
\end{figure*}

\subsection{Spatial Referring Expressions}
Referring expressions~\cite{krahmer2012computational} link natural language with visual perception~\cite{kazemzadeh2014referitgame,lin2015microsoftcococommonobjects,johnson2016clevrdiagnosticdatasetcompositional,yu2016modeling,yu2017joint}, enabling the description and identification of objects within images. These expressions often involve complex lexical structures; incorporating actions, object attributes, and spatial relationships, such as in the sentence, ``The girl eating ice cream to the left of the yellow table." Thus, to perform well on these benchmarks, a trained model does not necessarily need to fully understand the spatial relationships.

Spatial referring expressions, a specialized subset of referring expressions, emphasize the locations and spatial relationships of objects relative to nearby landmarks. Modeling spatial referring expressions presents unique challenges due to the need for contextual understanding. While appearance-based descriptions, such as color, shape, or object class~\cite{viethen2022generation}, focus solely on the attributes of the target object, spatial descriptions require comprehending the relationship between the landmark object and the target object. Additionally, spatial referring expressions are often perspective-dependent, adding another layer of complexity. By incorporating spatial context, these expressions enhance the clarity and precision of object identification, which is particularly important in scenes with multiple similar objects. Existing dataset~\cite{liu2021refer,liu2023visualspatialreasoning}, either lack comprehensive annotations for referred landmark objects, limit expressions by using ground truth object classnames from the image dataset, or contain too few landmark objects. In contrast, our work introduces complex spatial referring expressions with complete annotations.

\section{Methodology}
\label{sec:methodology}
\subsection{Dataset}
\begin{table*}[h!]
    \centering
    \resizebox{\textwidth}{!}{%
    \begin{tabular}{lccccc}
        \toprule
        \textbf{dataset} & \textbf{\# of annotations} & \textbf{\# of target objects} & \textbf{\# of objects per caption w/ image annotation} & \textbf{Annotated on} & \textbf{avgerage caption length} \\
        \midrule
        ScanRefer\cite{chen2020scanrefer3dobjectlocalization} & 51,583 & 11,046 & 1 & 704 scenes & 20.3 \\
        Nr3D\cite{achlioptas2020referit3d} & 41,503 & 5,879 & 1 & 642 scenes & 11.4 \\
        REVERIE\cite{qi2020reverieremoteembodiedvisual} & 21,702 & 4,140 & 1 & 90 scenes & 18.0 \\
        SUNRefer\cite{liu2021refer} & 38,495 & 7,699 & 1 & 7,699 RGBD images & 16.3 \\
        SUN-Spot\cite{mauceri2019sun} & 7,990 & 3,245 & 1 & 1,948 RGBD images & 14.1 \\
        \textbf{SUN-Spot v2.0 (expert)} & 7,990 & 3,245 & 2.91 & 1,948 RGBD images & 14.1 \\
        \textbf{SUN-Spot v2.0(machine-gen)} & 93,063 & 87,724 & 2.52 & 10,
        333 RGBD images & 10.0 \\
        \bottomrule
    \end{tabular}%
    }
    \caption{Comparison of SUN-Spot v2.0 and similar datasets.}
    \label{tab:compare_datasets}
\end{table*}

\begin{table}[h]
    \centering
    \begin{tabular}{l r r }
        \toprule
        Dataset & Expert &Machine-generated\\
        \# of descriptions & 7,990 &93,063\\
        \# of images & 1,948 &10,333\\
        \# of objects & 3,245 &87,724\\
        \# of objects per scene &20.26 &19.06\\
        \# of descriptions per image & 3.98 & 9.01\\    
        Size of vocabulary &2309 & 4484\\
        \bottomrule
    \end{tabular}
    \caption{SUN-Spot v2.0 dataset statistics}
    \label{tab:sunspot_stat}
\end{table}
\subsubsection{SUN-Spot v2.0 Expert}
The SUN-Spot v2.0 dataset extends the SUN RGB-D dataset~\cite{song2015sun}, which is comprised of over 10,000 RGB-D images of indoor scenes with 2D object segmentation and 3D object bounding boxes. SUN-Spot v2.0 Expert focuses on a subset of 1948 images labeled with 7987 spatial referring expressions (REs), averaging 2.6 spatial prepositions per expression. 

The annotation process consisted of two stages. Stage one introduced the SUN-Spot dataset, where human annotators provided spatial referring expressions (REs) with prompts designed to maximize spatial preposition usage. Unlike typical datasets that only include target object annotations, SUN-Spot aimed to reduce ambiguity in object references. Stage two improved visual grounding by requiring annotators to click on objects corresponding to highlighted landmarks in the captions. We then processed each click to determine which pre-existing ground truth segmentation mask it corresponded to, based on the click's pixel location. When a click falls onto multiple overlapping segmentation masks, the best mask is selected based on the similarity between the ground truth object label and the highlighted landmark object mentioned. On average, responses from three different human annotators were collected for each landmark object in every caption. To ensure the quality of the dataset, we hired an additional expert annotator to perform a second-round review of each caption and its landmark object annotation during the evaluation process.
\begin{figure*}[]
\centering
\includegraphics[width=1\textwidth]{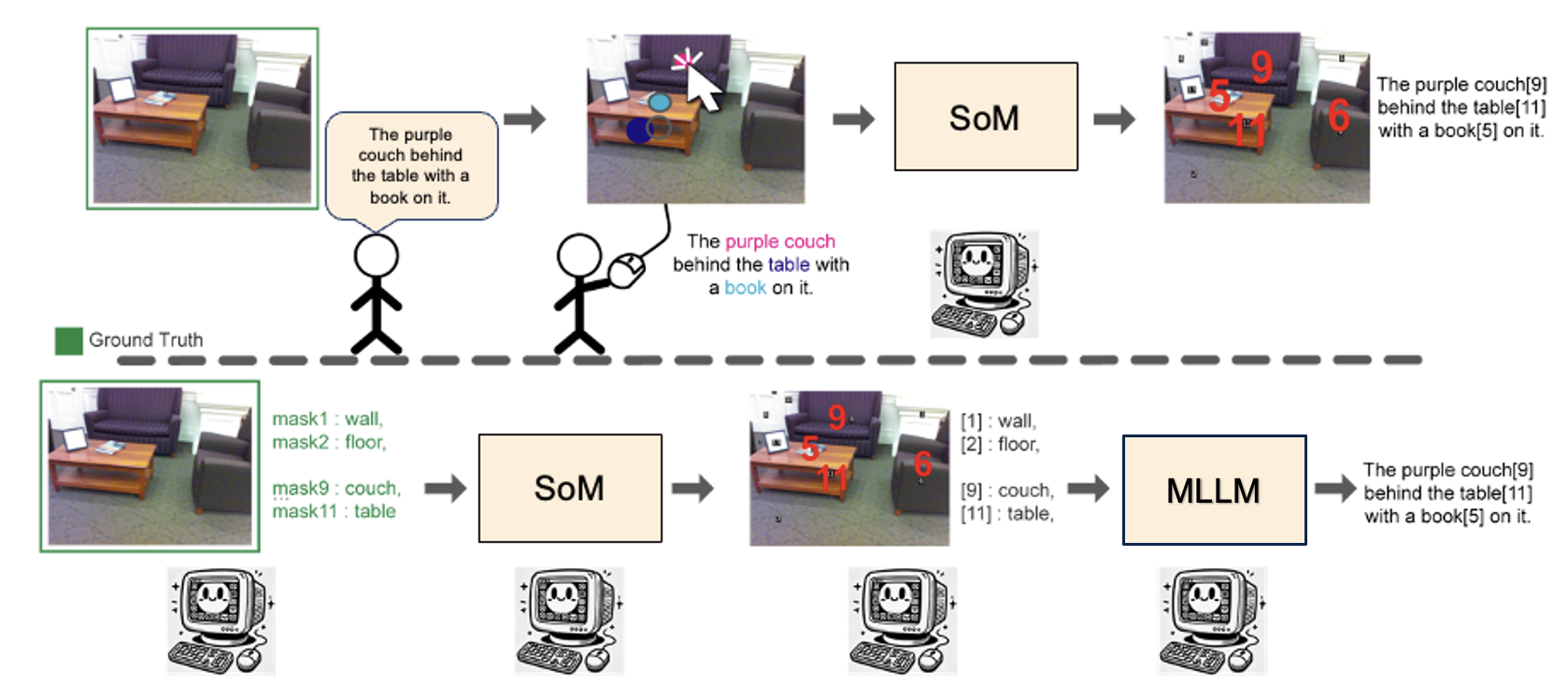}
\caption{ Caption generation and annotation process. Top: SUN-Spot V2.0 Expert dataset, Bottom: SUN-Spot V2.0 Machine-generated dataset}
\label{fig:caption_example}
\end{figure*}
The expert dataset provides higher quality and more complex spatial referring expressions compared to currently available datasets. We gathered and annotated a total of 7,987 spatial referring expressions. The dataset's high quality makes it an excellent resource for researchers seeking complex spatial referring expressions with strong alignment between objects in images and their mentions in captions. Consequently, SUN-Spot v2.0 Expert is designed to support further research in spatial language and scene understanding. See table \ref{tab:compare_datasets} for more detailed comparison to other similar datasets.

\subsubsection{SUN-Spot v2.0 Machine-generated}
\begin{figure}[]
\centering
\includegraphics[width=0.5\textwidth]{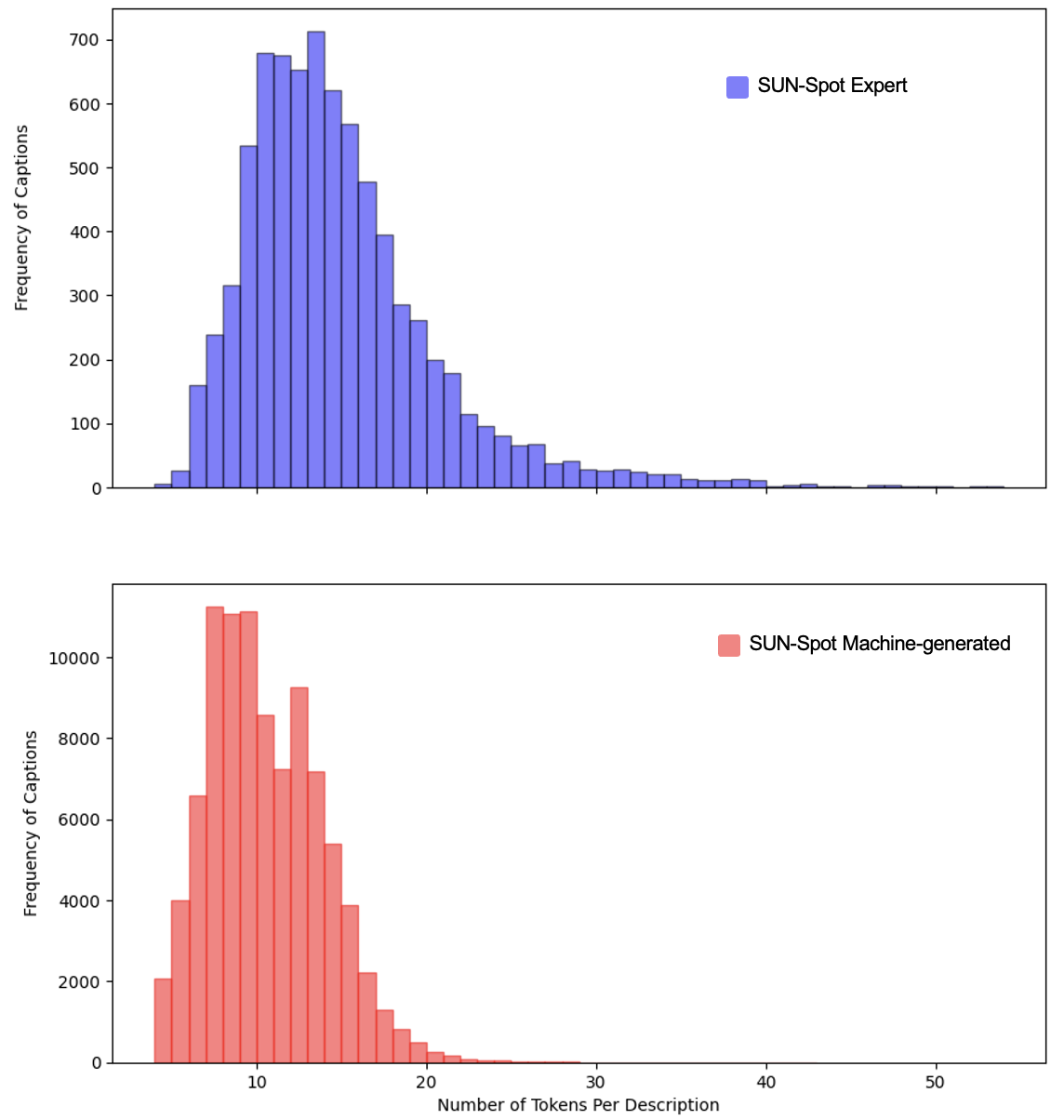}
\caption{Caption Length Comparison}
\label{fig:caption_length}
\end{figure}

To gather data more cost-effectively, we introduced an alternative approach for the SUN-Spot v2.0 Machine-generated dataset. First, we annotate images with SoM markers and input them, along with ground truth SoM labels, into an MLLM. The model then generates captions incorporating these markers, which we use to identify the corresponding object segmentation masks. See Figure \ref{fig:caption_example} for data collection details. Although large language models are not particularly strong at answering questions that involves spatial relationships~\cite{liu2023mmbench,kamath2023s}, they demonstrate good performance in caption generation. We can leverage these high-quality captions to train the models, enhancing their ability to answer questions more accurately and reducing their tendency to default to agreement with human input. However, this method has its limitations. The MLLM requires a high level of autonomy to produce high-quality captions. When prompts become more complex, such as requiring to include object attributes like color or size, or including two or more landmark objects in the caption, the quality of the captions can decline significantly. Furthermore, the MLLM depends on ground truth labels for objects to ensure accurate object mentions. Without these labels, relying solely on the MLLM’s object detection capabilities can lead to incorrect identifications. Additionally, the captions generated by the MLLM are noticeably shorter than those provided by human annotators. See Figure \ref{fig:caption_length} for more details. Despite those limitations, the method offers a viable and cost-effective alternative for data collection, particularly when the focus is on generating high-quality captions for training models. Moving forward, refining the balance between human annotation and machine-generated data will be key to further enhancing both the quality and efficiency of our dataset collection process. Table \ref{tab:sunspot_stat} provides a detailed comparison between SUN-Spot Expert and Machine-generated data. 

\subsubsection{SUNRefer}
Finetuning MLLM requires a relatively large amount of data, thus we also added SUNRefer~\cite{liu2021refer} data to our model training pipeline as additional data. SUNRefer contains 7699 RGB-D images from the SUNRGBD dataset and only one target object is chosen from each image. For each target object, five descriptions from
different annotators were collected in order to ensure linguistic diversity. SUNRefer includes a total of 38k spatial referring expressions.

\subsection{Set-of-Marks}
Set-of-Marks (SoM) prompting is a visual prompting method designed to enhance the visual grounding capabilities of large multimodal models like GPT-4V. SoM prompting avoids ambiguity by using abstract identifiers such as alphanumerics, masks, or boxes to mark regions within an image. Importantly, markers are consistently applied across different modalities: they are placed on the objects within the images to visually distinguish them, and the same markers are used in the captions, attached immediately after the corresponding object mentions. These markers serve as precise spatial references, enabling the alignment of language and visual input without over-relying on the semantic properties of the objects, thus minimizing ambiguity. 

According to 
~\cite{yang2023setofmarkpromptingunleashesextraordinary}, SoM offers enhanced capabilities to accurately map object references in language to specific image regions. This approach is crucial because it addresses situations where semantic assumptions could lead to inaccuracies. For example, if a cup is placed on the floor next to a table, relying heavily on semantic information might incorrectly lead the model to assume the cup is on the table. SoM helps weaken the model’s reliance on such semantic cues, thereby improving its comprehension of actual spatial positions. This adjustment allows Spatial-LLaVA to more accurately reflect real-world spatial relationships in visual content, enhancing the model’s utility in complex scenarios where accurate spatial understanding is key. See Figure \ref{fig:system_example} for an example of how we apply SoM to an image in our dataset.

\subsection{Conversation Generation}
\label{ssec:conversation_gen}

Despite the abundance of visual question answering datasets available online, they often contain minimal or no spatial referring expressions. In line with the original LLaVA work, we employ the language-only LLM, GPT-4o, to generate conversations composed of a series of questions and answers that focus on the spatial relationship between objects mentioned in the captions. The input to GPT-4o comprises two elements: firstly, captions from the SUN-Spot v2.0 Expert and SUNRefer dataset that describe the locations of one or multiple targets using spatial referring expressions. Secondly, SoM markers were added after each object mention for both target and landmark objects providing additional spatial context. We employ a few-shot learning approach in our conversation generation pipeline, which involves manually constructing conversations for a few images and using these as initial inputs to GPT-4o before introducing real examples.  This method, while similar to other conversation generation methods, is distinguished by our specific use of prompts that rigorously adhere to spatial referring expressions. This ensures that the questions asked are focused on exploring spatial relationships, rather than broadly generated content by GPT-4o. 
We collected a total of 75k question-answer pairs using the SUN-Spot v2.0 Expert and the SUNRefer dataset. One illustrative example of this approach can be seen in Figure \ref{fig:system_example}.
%\textbf{GPT-4o &  65.57 & 64.92 & 66.06 & 64.66}
\begin{table*}[h!]
    \centering
    \begin{tabular}{|l|c|c|c|c|}
        \hline
        \textbf{Model} & \textbf{Accuracy (\%)} & \textbf{Precision (\%)} & \textbf{Recall (\%)} & \textbf{F1 Score (\%)} \\
        \hline
        BLIP2\cite{li2023blip2bootstrappinglanguageimagepretraining} & 37.55 & 48.04 & 49.57 & 30.87 \\
        InstructBLIP\cite{dai2023instructblipgeneralpurposevisionlanguagemodels} & 48.51 & 28.68 & 38.22 & 32.74 \\
        BLIP\cite{li2022blipbootstrappinglanguageimagepretraining} & 49.65 & 51.09 & 51.16 & 49.31 \\
        ALBEF\cite{li2021alignfusevisionlanguage} & 61.48 & 48.32 & 49.58 & 42.53 \\
        Qwen-VL-Chat\cite{bai2023qwenvlversatilevisionlanguagemodel}&  63.13 & 56.83 & 53.19 & 50.17 \\
        PaliGemma\cite{beyer2024paligemmaversatile3bvlm} & 66.03 & 62.32 & 59.54 & 59.52  \\
        \hline
        LLaVA\_v1.5\_7b~\cite{liu2023LLaVA} & 64.17 & 59.87 & 57.93 & 57.84 \\
         LLaVA\_v1.5\_13b~\cite{liu2023LLaVA} & 66.22 & 62.67 & 58.79 & 58.39 \\

        \hline
         Spatial-LLaVA\_7b & 71.62 & 69.57 & 66.55 & 67.19 \\
        Spatial-LLaVA\_13b &  \textbf{76.13} &  \textbf{75.83} &  \textbf{76.13} &  \textbf{75.92} \\
        \hline
    \end{tabular}
    \caption{Results, SUN-Spot v2.0 Expert Benchmark}
    \label{tab:metrics_SUN-Spot}
\end{table*}

\subsection{Model}
As previously discussed, the pre-processing of each image and its corresponding caption utilizes %Set-of-Marks (SoM)
SoM prompting. While our approach incorporates SoM, it aligns with methodologies from ViP-LLaVA~\cite{cai2024vipLLaVAmakinglargemultimodal} and LLaVA by employing a pre-trained CLIP visual encoder to process each input image, thereby outputting the visual features. Unlike previous efforts that capture object-level image features, our approach embeds SoM markers as tiny digits at the center of each object.

The image features are then passed to a projection layer—the only component with trainable parameters in this pipeline except MLLM—which converts the image features into language embedding tokens. These tokens are subsequently concatenated with language embeddings extracted from the conversation data, forming a comprehensive representation that seamlessly bridges visual and language input.

Spatial-LLaVA is trained in two stages: the first stage focuses on aligning visual features with language features. Here, we utilize the LLaVA pre-trained weights, which are pre-trained on the LLaVA v1.5 instruction dataset alongside their own dataset. In the second stage, the weights of the visual encoder remain frozen, and only the projection parameters are updated based on our conversation data. LLaVA v1.5 is trained on multi-turn conversation data, allowing the model to access all previous questions and answers when predicting the answer for the current question. Given that questions in our dataset often focus on object relationships, knowing the answer to one question can sometimes reveal the answer to another. To prevent the model from relying on such inferences and to ensure it treats each question independently, we chose to fine-tune it on single-turn conversation data. This method helps the model generate more independent and unbiased predictions for each question.

This two-stage approach is adopted due to the high computational costs and large datasets required for initial training in visual and language alignment. Given the limited dataset available for spatial referring expressions, fine-tuning the pre-trained model with our domain-specific data proves to be an effective strategy. This method leverages the established benefits of using pre-trained weights, significantly improving model performance by allowing quick adaptation to task-specific requirements without the need for extensive retraining from scratch.

\section{Experiments}
\label{sec:result}
In this section, we will discuss the training setup and perform a comparative analysis of Spatial-LLaVA against %various 
a selection of state-of-the-art LLMs. %open-sourced LLMs.
% GPT-4o & 33.00 & 32.46 & 33.36 & 32.35 \\
\begin{table*}[h!]
    \centering
    \begin{tabular}{|l|c|c|c|c|}
        \hline
        \textbf{Model} & \textbf{Accuracy (\%)} & \textbf{Precision (\%)} & \textbf{Recall (\%)} & \textbf{F1 Score (\%)} \\
        \hline
        
        InstructBLIP\cite{dai2023instructblipgeneralpurposevisionlanguagemodels} & 52.05 & 53.99 & 50.71 & 38.59 \\
        BLIP\cite{li2022blipbootstrappinglanguageimagepretraining} & 45.25 & 39.70 & 44.27 & 37.72 \\
        ALBEF\cite{li2021alignfusevisionlanguage} & 51.47 & 50.74 & 50.01 & 34.43 \\
        Qwen-VL-Chat\cite{bai2023qwenvlversatilevisionlanguagemodel}& 53.44 & 58.30 & 52.17 & 41.98\\
        PaliGemma\cite{beyer2024paligemmaversatile3bvlm}& 53.58 & 54.86 & 52.40 & 46.11 \\
        \hline
         LLaVA\_v1.5\_7b~\cite{liu2023LLaVA} & 52.95 & 57.99 & 51.63 & 40.27 \\
        LLaVA\_v1.5\_13b~\cite{liu2023LLaVA} & 52.37 & 54.79 & 51.08 & 39.94 \\
        \hline
         Spatial-LLaVA\_7b & 53.60 & 54.77 & 52.61 & 47.08 \\
        Spatial-LLaVA\_13b &  \textbf{56.14} &  \textbf{60.19} &  \textbf{55.10} &  \textbf{49.26} \\
        \hline
    \end{tabular}
    \caption{Results, Visual Spatial Reasoning Benchmark}
    \label{tab:metrics_VSR}
\end{table*}

\begin{figure*}[]
\centering
\includegraphics[width=1\textwidth]{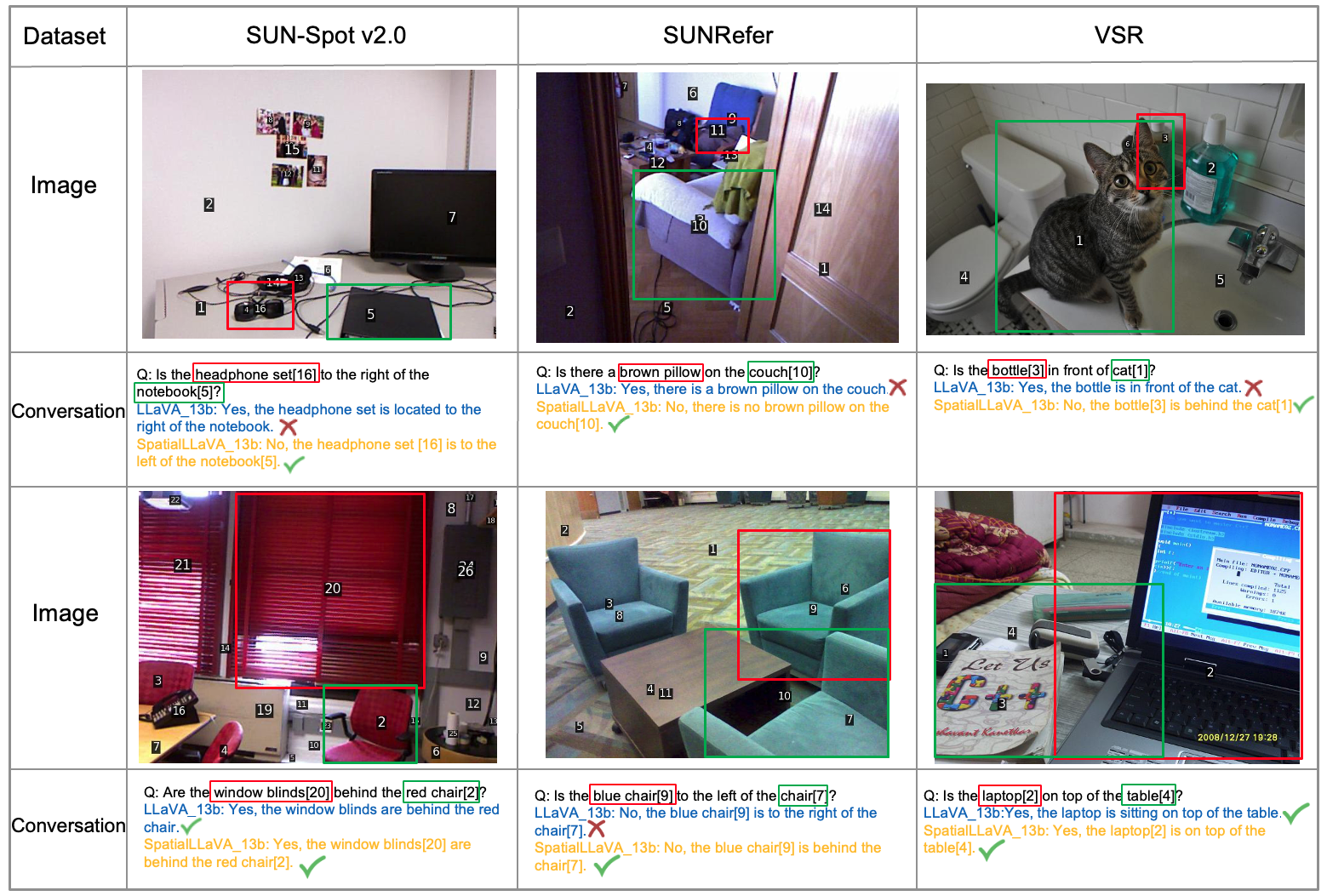}
\caption{Comparison of Spatial-LLaVA and LLaVA Across Datasets: SUN-Spot v2.0, SUNRefer, and VSR}
\label{fig:more_examples}
\end{figure*}

\subsection{Training Setup}
We selected Vicuna-v1.5~\cite{vicuna2023} as our language tokenizer to encode the conversation and the pre-trained CLIP visual encoder ViT-L/14 to transform the image into a set of tokens. These tokens are then projected into the same space as the language tokens using a 2-layer MLP. Subsequently, the concatenated features are fed into the large language model, Vicuna, for fine-tuning. During this process, the parameters of the MLP projector and the LLM are updated, while the weights of the vision encoder and language tokenizer remain frozen.

\subsection{Training and Data}
During the initial stage of training, the model was pre-trained using a filtered CC-595K subset for one epoch with a learning rate of 2e-3 and a batch size of 128. In total, we generated 75k question-answer pairs using our SUN-Spot Expert and SUNRefer dataset, and then utilized this spatial conversation data to fine-tune the pre-trained weights. The model was fine-tuned for three epochs with a learning rate of 2e-6 and a batch size of 128. The fine-tuning process took approximately 2 hours on 8 A100 GPUs.

\subsection{Baseline Models}
We compare Spatial-LLaVA against eight state-of-the-art MLLMs, including BLIP2\cite{li2023blip2bootstrappinglanguageimagepretraining}, InstructBLIP\cite{dai2023instructblipgeneralpurposevisionlanguagemodels}, BLIP\cite{li2022blipbootstrappinglanguageimagepretraining}, ALBEF\cite{li2021alignfusevisionlanguage}, Qwen-VL-Chat\cite{bai2023qwenvlversatilevisionlanguagemodel}, PaliGemma\cite{beyer2024paligemmaversatile3bvlm}, LLaVA 7b
, and LLaVA 13b~\cite{liu2023LLaVA}. These models were chosen for their excellence in various vision-language tasks such as image captioning, visual question answering, and cross-modal retrieval, enabling  a thorough comparison of Spatial-LLaVA's performance against leading models. For BLIP, we use the VQA checkpoint fine-tuned on VQAv2. BLIP2 is evaluated in its largest configuration, with ViT-g as the frozen image encoder and FlanT5-XXL as the frozen LLM. InstructBLIP, an instruction-tuned extension of BLIP-2, is tested with frozen Vicuna\_7b weights. ALBEF is used with its model pre-trained on 14M images and fine-tuned on VQAv2. PaliGemma is assessed using the paligemma-3b-mix-224 weights, while Qwen-VL uses fine-tuned weights from Qwen-VL-Chat.

\subsection{Evaluation on the Spatial Benchmark Dataset}

\subsubsection{Evaluation on SUN-Spot V2.0 Expert and SUNRefer}
When generating conversations for training, three main types of questions are asked as in Section \ref{ssec:conversation_gen}. However, evaluating these questions is challenging due to the high similarity between correct and incorrect answers. For example, the sentences ``The yellow cup is on the left side of the phone'' and ``The yellow cup is on the right side of the phone'' have a similarity of 95\% when encoded using the BERT encoder. To address this issue, we simplified the questions to binary answers, asking the fine-tuned model to determine whether the spatial relationship exists between the target object and the landmark objects. 

The evaluation is conducted using four key metrics: Accuracy, Precision, Recall, and F1 Score, ensuring a thorough assessment of the model’s ability to classify spatial relationships. Accuracy reflects overall correctness in determining whether a spatial relationship exists. Precision indicates how often predicted relationships are correct, reducing false positives. Recall measures the model’s ability to detect all true spatial relationships, minimizing false negatives. F1 Score, as a balance of Precision and Recall, is particularly important given the subtle differences between correct and incorrect spatial descriptions. These metrics together capture the model’s effectiveness in understanding and predicting spatial relationships.

The MLLMs are tested using the SUN-Spot v2.0 Expert and SUNRefer test datasets, which comprise 10\% of the total data and are held out separately from the training set. As demonstrated in Table \ref{tab:metrics_SUN-Spot},Spatial-LLaVA achieves state-of-the-art performance, significantly surpassing the original LLaVA by 7.5\% for the 7b model and 9.91\% for the 13b model. To further understand these advancements, we conducted an in-depth analysis to determine the specific spatial relationships where Spatial-LLaVA has shown improvement compared to models with relatively close performance. Our findings reveal that Spatial-LLaVA excels in several key spatial relationships, including ``above'', ``below'', ''under'', ``in front of'', ``between'', ``behind'', ``left'', ``right'', and ``on top of''. The improvements in Spatial-LLaVA are particularly clear in its ability to interpret directional relationships like "above" and "below," which are key for spatial orientation and navigation. The model also excels in understanding relative positioning (``in front of'', ``behind'', ``between'') and hierarchical relationships (``under'', ``on top of''), enhancing its accuracy in predicting object placement.

\subsubsection{Evaluation Results on Visual Spatial Reasoning}
Table \ref{tab:metrics_VSR} presents the comparison with MLLMs using the Visual-Spatial Reasoning (VSR) ~\cite{liu2023visualspatialreasoning} test set, which contains a total of 1,222 questions. We performed some prepossessing on the questions, originally constructed as caption-label pairs with ``True" and ``False" labels.  To streamline the process, we reconstructed the captions into questions, eliminating the need for additional prompting to explain tasks, with answers now being ``Yes'' or ``No''. For example, the caption ``The cup on top of the table'' with the label ``True'' was transformed into ``Is the cup on top of the table?" with the answer ``Yes.'' Additionally, we incorporated SoM prompting into the questions to ensure that unique objects are clearly referenced. All the questions were answered as a zero-shot question-answering task. As shown in the table\ref{tab:metrics_VSR}, Spatial-LLaVA outperforms other models on the VSR dataset, demonstrating that our collected dataset features high-quality annotations and is suitable for various general downstream tasks. We also analyzed the spatial relationships present in the VSR dataset, which emphasize different relationships compared to those in our dataset, resulting in limited overlap between the two sets. See Figure \ref{fig:example_3_datasets} for qualitative comparison. More detailed discussion is provided in the appendix. 

\section{Conclusion and Future Work}
In this work, we introduced the SUN-Spot v2.0 dataset - the first RGB-D dataset to include both spatial referring expressions and detailed annotations for target and landmark objects. We presented Spatial-LLaVA, a fine-tuned MLLM specifically designed to understand and predict spatial relationships between objects in images using Set-of-Marks prompting. Spatial-LLaVA outperforms existing models in spatial reasoning tasks, underscoring the importance of precise spatial references and reducing the reliance on semantic object labels. We hope this approach will inspire further research in developing more MLLMs that are capable of handling complex tasks in real-world scenarios. Building on these advancements, future research could explore the application of Spatial-LLaVA and similar models in visual language navigation task. Specifically, integrating Spatial-LLaVA with robotic systems could enhance their ability to interpret and act upon spatial instructions provided in natural language, thereby improving autonomous navigation and interaction in complex environments. Further studies could investigate real-time navigation scenarios and expand the dataset to include more diverse environments and objects, which could enhance model robustness and lead to the development of more capable robotic systems.

\bibliographystyle{plain}
\bibliography{references}

\begin{thebibliography}{10}

\bibitem{achlioptas2020referit3d}
Panos Achlioptas, Ahmed Abdelreheem, Fei Xia, Mohamed Elhoseiny, and Leonidas Guibas.
\newblock Referit3d: Neural listeners for fine-grained 3d object identification in real-world scenes.
\newblock In {\em Computer Vision--ECCV 2020: 16th European Conference, Glasgow, UK, August 23--28, 2020, Proceedings, Part I 16}, pages 422--440. Springer, 2020.

\bibitem{bai2023qwenvlversatilevisionlanguagemodel}
Jinze Bai, Shuai Bai, Shusheng Yang, Shijie Wang, Sinan Tan, Peng Wang, Junyang Lin, Chang Zhou, and Jingren Zhou.
\newblock Qwen-vl: A versatile vision-language model for understanding, localization, text reading, and beyond, 2023.

\bibitem{beyer2024paligemmaversatile3bvlm}
Lucas Beyer, Andreas Steiner, André~Susano Pinto, Alexander Kolesnikov, Xiao Wang, Daniel Salz, Maxim Neumann, Ibrahim Alabdulmohsin, Michael Tschannen, Emanuele Bugliarello, Thomas Unterthiner, Daniel Keysers, Skanda Koppula, Fangyu Liu, Adam Grycner, Alexey Gritsenko, Neil Houlsby, Manoj Kumar, Keran Rong, Julian Eisenschlos, Rishabh Kabra, Matthias Bauer, Matko Bošnjak, Xi~Chen, Matthias Minderer, Paul Voigtlaender, Ioana Bica, Ivana Balazevic, Joan Puigcerver, Pinelopi Papalampidi, Olivier Henaff, Xi~Xiong, Radu Soricut, Jeremiah Harmsen, and Xiaohua Zhai.
\newblock Paligemma: A versatile 3b vlm for transfer, 2024.

\bibitem{cai2024vipLLaVAmakinglargemultimodal}
Mu~Cai, Haotian Liu, Dennis Park, Siva~Karthik Mustikovela, Gregory~P. Meyer, Yuning Chai, and Yong~Jae Lee.
\newblock Vip-llava: Making large multimodal models understand arbitrary visual prompts, 2024.

\bibitem{carion_2020_endtoend}
Nicolas Carion, Francisco Massa, Gabriel Synnaeve, Nicolas Usunier, Alexander Kirillov, and Sergey Zagoruyko.
\newblock End-to-end object detection with transformers.
\newblock {\em arXiv:2005.12872 [cs]}, 05 2020.

\bibitem{chen2020scanrefer3dobjectlocalization}
Dave~Zhenyu Chen, Angel~X. Chang, and Matthias Nießner.
\newblock Scanrefer: 3d object localization in rgb-d scans using natural language, 2020.

\bibitem{mehdicherti_2023_reproducible}
Mehdi Cherti, Romain Beaumont, Ross Wightman, Mitchell Wortsman, Gabriel Ilharco, Cade Gordon, Christoph Schuhmann, Ludwig Schmidt, and Jenia Jitsev.
\newblock Reproducible scaling laws for contrastive language-image learning.
\newblock 06 2023.

\bibitem{vicuna2023}
Wei-Lin Chiang, Zhuohan Li, Zi~Lin, Ying Sheng, Zhanghao Wu, Hao Zhang, Lianmin Zheng, Siyuan Zhuang, Yonghao Zhuang, Joseph~E. Gonzalez, Ion Stoica, and Eric~P. Xing.
\newblock Vicuna: An open-source chatbot impressing gpt-4 with 90\%* chatgpt quality, March 2023.

\bibitem{chowdhery_2022_palm}
Aakanksha Chowdhery, Sharan Narang, and et~al. Devlin, Jacob.
\newblock Palm: Scaling language modeling with pathways.
\newblock {\em arXiv:2204.02311 [cs]}, 04 2022.

\bibitem{chung_2022_scaling}
Hyung~Won Chung, Le~Hou, and et~al.0 Longpre, Shayne.
\newblock Scaling instruction-finetuned language models.
\newblock {\em arXiv:2210.11416 [cs]}, 10 2022.

\bibitem{dai2023instructblipgeneralpurposevisionlanguagemodels}
Wenliang Dai, Junnan Li, Dongxu Li, Anthony Meng~Huat Tiong, Junqi Zhao, Weisheng Wang, Boyang Li, Pascale Fung, and Steven Hoi.
\newblock Instructblip: Towards general-purpose vision-language models with instruction tuning, 2023.

\bibitem{dettmers2023qloraefficientfinetuningquantized}
Tim Dettmers, Artidoro Pagnoni, Ari Holtzman, and Luke Zettlemoyer.
\newblock Qlora: Efficient finetuning of quantized llms, 2023.

\bibitem{devlin_2018_bert}
Jacob Devlin, Ming-Wei Chang, Kenton Lee, and Kristina Toutanova.
\newblock Bert: Pre-training of deep bidirectional transformers for language understanding, 10 2018.

\bibitem{fang_2022_eva}
Yuxin Fang, Wen Wang, Binhui Xie, Quan Sun, Ledell Wu, Xinggang Wang, Tiejun Huang, Xinlong Wang, and Yue Cao.
\newblock Eva: Exploring the limits of masked visual representation learning at scale, 12 2022.

\bibitem{geminiteam2024geminifamilyhighlycapable}
{Gemini Team}, Rohan Anil, and et~al. Sebastian~Borgeaud.
\newblock Gemini: A family of highly capable multimodal models, 2024.

\bibitem{heakl_6th}
Ahmed Heakl, Youssef Zaghloul, Mennatullah Ali, Rania Hossam, and Walid Gomaa.
\newblock Arzen-llm: Code-switched egyptian arabic-english translation and speech recognition using llms.
\newblock In {\em 6th International Conference on AI in Computational Linguistics}, 2024.

\bibitem{johnson2016clevrdiagnosticdatasetcompositional}
Justin Johnson, Bharath Hariharan, Laurens van~der Maaten, Li~Fei-Fei, C.~Lawrence Zitnick, and Ross Girshick.
\newblock Clevr: A diagnostic dataset for compositional language and elementary visual reasoning, 2016.

\bibitem{kamath2023s}
Amita Kamath, Jack Hessel, and Kai-Wei Chang.
\newblock What's ``up" with vision-language models? investigating their struggle with spatial reasoning.
\newblock {\em arXiv preprint arXiv:2310.19785}, 2023.

\bibitem{kazemzadeh2014referitgame}
Sahar Kazemzadeh, Vicente Ordonez, Mark Matten, and Tamara Berg.
\newblock Referitgame: Referring to objects in photographs of natural scenes.
\newblock In {\em Proceedings of the 2014 conference on empirical methods in natural language processing (EMNLP)}, pages 787--798, 2014.

\bibitem{krahmer2012computational}
Emiel Krahmer and Kees Van~Deemter.
\newblock Computational generation of referring expressions: A survey.
\newblock {\em Computational Linguistics}, 38(1):173--218, 2012.

\bibitem{labruna_2024_when}
Tiziano Labruna, Jon~Ander Campos, and Gorka Azkune.
\newblock When to retrieve: Teaching llms to utilize information retrieval effectively, 05 2024.

\bibitem{li2023blip2bootstrappinglanguageimagepretraining}
Junnan Li, Dongxu Li, Silvio Savarese, and Steven Hoi.
\newblock Blip-2: Bootstrapping language-image pre-training with frozen image encoders and large language models, 2023.

\bibitem{li2022blipbootstrappinglanguageimagepretraining}
Junnan Li, Dongxu Li, Caiming Xiong, and Steven Hoi.
\newblock Blip: Bootstrapping language-image pre-training for unified vision-language understanding and generation, 2022.

\bibitem{li2021alignfusevisionlanguage}
Junnan Li, Ramprasaath~R. Selvaraju, Akhilesh~Deepak Gotmare, Shafiq Joty, Caiming Xiong, and Steven Hoi.
\newblock Align before fuse: Vision and language representation learning with momentum distillation, 2021.

\bibitem{lin2015microsoftcococommonobjects}
Tsung-Yi Lin, Michael Maire, Serge Belongie, Lubomir Bourdev, Ross Girshick, James Hays, Pietro Perona, Deva Ramanan, C.~Lawrence Zitnick, and Piotr Dollár.
\newblock Microsoft coco: Common objects in context, 2015.

\bibitem{liu2023visualspatialreasoning}
Fangyu Liu, Guy Emerson, and Nigel Collier.
\newblock Visual spatial reasoning, 2023.

\bibitem{liu2021refer}
Haolin Liu, Anran Lin, Xiaoguang Han, Lei Yang, Yizhou Yu, and Shuguang Cui.
\newblock Refer-it-in-rgbd: A bottom-up approach for 3d visual grounding in rgbd images.
\newblock In {\em Proceedings of the IEEE/CVF Conference on Computer Vision and Pattern Recognition}, pages 6032--6041, 2021.

\bibitem{liu2023improvedLLaVA}
Haotian Liu, Chunyuan Li, Yuheng Li, and Yong~Jae Lee.
\newblock Improved baselines with visual instruction tuning, 2023.

\bibitem{liu2023LLaVA}
Haotian Liu, Chunyuan Li, Qingyang Wu, and Yong~Jae Lee.
\newblock Visual instruction tuning.
\newblock In {\em NeurIPS}, 2023.

\bibitem{liu2019clevr}
Runtao Liu, Chenxi Liu, Yutong Bai, and Alan~L Yuille.
\newblock Clevr-ref+: Diagnosing visual reasoning with referring expressions.
\newblock In {\em Proceedings of the IEEE/CVF conference on computer vision and pattern recognition}, pages 4185--4194, 2019.

\bibitem{liu_2019_roberta}
Yinhan Liu, Myle Ott, Naman Goyal, Jingfei Du, Mandar Joshi, Danqi Chen, Omer Levy, Mike Lewis, Luke Zettlemoyer, Veselin Stoyanov, and Paul Allen.
\newblock Roberta: A robustly optimized bert pretraining approach, 2019.

\bibitem{liu2023mmbench}
Yuan Liu, Haodong Duan, Yuanhan Zhang, Bo~Li, Songyang Zhang, Wangbo Zhao, Yike Yuan, Jiaqi Wang, Conghui He, Ziwei Liu, et~al.
\newblock Mmbench: Is your multi-modal model an all-around player?
\newblock {\em arXiv preprint arXiv:2307.06281}, 2023.

\bibitem{ma_an}
Ziyang Ma, Guanrou Yang, Yifan Yang, Zhifu Gao, Jiaming Wang, Zhihao Du, Yu~Fan, Qian Chen, Siqi Zheng, Shiliang Zhang, and Xie Chen.
\newblock An embarrassingly simple approach for llm with strong asr capacity.

\bibitem{mao2016generationcomprehensionunambiguousobject}
Junhua Mao, Jonathan Huang, Alexander Toshev, Oana Camburu, Alan Yuille, and Kevin Murphy.
\newblock Generation and comprehension of unambiguous object descriptions, 2016.

\bibitem{mauceri2019sun}
Cecilia Mauceri, Martha Palmer, and Christoffer Heckman.
\newblock Sun-spot: An rgb-d dataset with spatial referring expressions.
\newblock In {\em Proceedings of the IEEE/CVF International Conference on Computer Vision Workshops}, pages 0--0, 2019.

\bibitem{moslem_2023_domain}
Yasmin Moslem, Gianfranco Romani, Mahdi Molaei, Rejwanul Haque, John Kelleher, and Andy Way.
\newblock Domain terminology integration into machine translation: Leveraging large language models, 2023.

\bibitem{journaloflatexclass_2015_mmger}
Bingshen Mu, Yangze Li, Qijie Shao, Kun Wei, Xucheng Wan, Naijun Zheng, Huan Zhou, and Lei Xie.
\newblock Mmger: Multi-modal and multi-granularity generative error correction with llm for joint accent and speech recognition.
\newblock {\em arXiv preprint arXiv:2405.03152}, 2024.

\bibitem{openai_2023_gpt4}
OpenAI.
\newblock Gpt-4 technical report.
\newblock \url{https://arxiv.org/pdf/2303.08774v3}, 2023.
\newblock Accessed: 2024-07-29.

\bibitem{qi2020reverieremoteembodiedvisual}
Yuankai Qi, Qi~Wu, Peter Anderson, Xin Wang, William~Yang Wang, Chunhua Shen, and Anton van~den Hengel.
\newblock Reverie: Remote embodied visual referring expression in real indoor environments, 2020.

\bibitem{radford_2021_learning}
Alec Radford, Jong~Wook Kim, Chris Hallacy, Aditya Ramesh, Gabriel Goh, Sandhini Agarwal, Girish Sastry, Amanda Askell, Pamela Mishkin, Jack Clark, Gretchen Krueger, and Ilya Sutskever.
\newblock Learning transferable visual models from natural language supervision.
\newblock {\em arXiv:2103.00020 [cs]}, 02 2021.

\bibitem{song2015sun}
Shuran Song, Samuel~P Lichtenberg, and Jianxiong Xiao.
\newblock Sun rgb-d: A rgb-d scene understanding benchmark suite.
\newblock In {\em Proceedings of the IEEE conference on computer vision and pattern recognition}, pages 567--576, 2015.

\bibitem{sun_2023_evaclip}
Quan Sun, Yuxin Fang, Ledell Wu, Xinlong Wang, and Yue Cao.
\newblock Eva-clip: Improved training techniques for clip at scale, 03 2023.

\bibitem{tang_selfretrieval}
Qiaoyu Tang, Jiawei Chen, Bowen Yu, Yaojie Lu, Cheng Fu, Haiyang Yu, Hongyu Lin, Fei Huang, Ben He, Xianpei Han, Le~Sun, and Yongbin Li.
\newblock Self-retrieval: Building an information retrieval system with one large language model.

\bibitem{touvron_2023_llama}
Hugo Touvron, Thibaut Lavril, Gautier Izacard, Xavier Martinet, Marie-Anne Lachaux, Timothée Lacroix, Baptiste Rozière, Naman Goyal, Eric Hambro, Faisal Azhar, Aurelien Rodriguez, Armand Joulin, Edouard Grave, and Guillaume Lample.
\newblock Llama: Open and efficient foundation language models.
\newblock {\em arXiv:2302.13971 [cs]}, 02 2023.

\bibitem{vaswani2023attentionneed}
Ashish Vaswani, Noam Shazeer, Niki Parmar, Jakob Uszkoreit, Llion Jones, Aidan~N. Gomez, Lukasz Kaiser, and Illia Polosukhin.
\newblock Attention is all you need, 2023.

\bibitem{viethen2022generation}
Henriette Anna~Elisabeth Viethen.
\newblock {\em The generation of natural descriptions: corpus-based investigations of referring expressions in visual domains}.
\newblock PhD thesis, Macquarie University, 2022.

\bibitem{yang2023setofmarkpromptingunleashesextraordinary}
Jianwei Yang, Hao Zhang, Feng Li, Xueyan Zou, Chunyuan Li, and Jianfeng Gao.
\newblock Set-of-mark prompting unleashes extraordinary visual grounding in gpt-4v, 2023.

\bibitem{yang_2019_xlnet}
Zhilin Yang, Zihang Dai, Yiming Yang, Jaime Carbonell, Ruslan Salakhutdinov, and Quoc~V Le.
\newblock Xlnet: Generalized autoregressive pretraining for language understanding, 2019.

\bibitem{yu2016modeling}
Licheng Yu, Patrick Poirson, Shan Yang, Alexander~C Berg, and Tamara~L Berg.
\newblock Modeling context in referring expressions.
\newblock In {\em Computer Vision--ECCV 2016: 14th European Conference, Amsterdam, The Netherlands, October 11-14, 2016, Proceedings, Part II 14}, pages 69--85. Springer, 2016.

\bibitem{yu2017joint}
Licheng Yu, Hao Tan, Mohit Bansal, and Tamara~L Berg.
\newblock A joint speaker-listener-reinforcer model for referring expressions.
\newblock In {\em Proceedings of the IEEE conference on computer vision and pattern recognition}, pages 7282--7290, 2017.

\bibitem{zhang_2023_prompting}
Biao Zhang, Barry Haddow, and Alexandra Birch.
\newblock Prompting large language model for machine translation: A case study, 2023.

\bibitem{NEURIPS2023_9cb2a749}
Yuchen Zhuang, Yue Yu, Kuan Wang, Haotian Sun, and Chao Zhang.
\newblock Toolqa: A dataset for llm question answering with external tools.
\newblock In A.~Oh, T.~Naumann, A.~Globerson, K.~Saenko, M.~Hardt, and S.~Levine, editors, {\em Advances in Neural Information Processing Systems}, volume~36, pages 50117--50143. Curran Associates, Inc., 2023.

\end{thebibliography}

\end{document}